\documentclass{article}
\usepackage{spconf,amsmath,graphicx}

\usepackage{cite}
\usepackage{amsmath,amssymb,amsfonts}
\usepackage{mathtools}
\usepackage{commath}
\usepackage{algorithmic}
\usepackage{graphicx}
\usepackage{textcomp}
\usepackage{xcolor}

\usepackage{fixltx2e}

\usepackage{booktabs}
\usepackage{array}
\definecolor{mygreen}{rgb}{0, 0.56, 0}

\usepackage{lipsum}
\usepackage[hidelinks]{hyperref}

\usepackage[misc,geometry]{ifsym}

\usepackage{float}
\floatstyle{plaintop}
\restylefloat{table}

\usepackage{enumitem}
\setlist{nosep, leftmargin=14pt}

\usepackage{mwe} 

\newcommand\extrafootertext[1]{%
    \bgroup
    \renewcommand\thefootnote{\fnsymbol{footnote}}%
    \renewcommand\thempfootnote{\fnsymbol{mpfootnote}}%
    \footnotetext[0]{#1}%
    \egroup
}

\title{SegReg: Segmenting OARs by Registering MR Images and CT Annotations}
\name{
\begin{tabular}{c}
    Zeyu Zhang$^{1,2,*}$, Xuyin Qi$^{3}$, Bowen Zhang$^{1}$, Biao Wu$^{1}$, Hien Le$^{4}$, Bora Jeong$^{5,6}$, \\
    Zhibin Liao$^{1}$, Yunxiang Liu$^{1}$, Johan Verjans$^{1}$, Minh-Son To$^{1,7}$, Richard Hartley$^{2, \text{\Letter}}$
\end{tabular}
}
\address{\fontsize{10}{4}\selectfont $^{1}$ Australian Institute for Machine Learning, The University of Adelaide 
$^{2}$ The Australian National University \\
\fontsize{10}{4}\selectfont $^{3}$ School of Computer and Mathematical Sciences, The University of Adelaide \\
\fontsize{10}{4}\selectfont $^{4}$Australian Bragg Centre for Proton Therapy and Research, South Australian Health and Medical Research Institute \\
\fontsize{10}{4}\selectfont $^{5}$ JBI, The University of Adelaide
\fontsize{10}{4}\selectfont $^{6}$ Department of Otolaryngology, Modbury Hospital\\
\fontsize{10}{4}\selectfont $^{7}$ Flinders Health and Medical Research Institute, Flinders University 
}

\begin{document}
%
\maketitle
\begin{abstract}
Organ at risk (OAR) segmentation is a critical process in radiotherapy treatment planning such as head and neck tumors. Nevertheless, in clinical practice, radiation oncologists predominantly perform OAR segmentations manually on CT scans. This manual process is highly time-consuming and expensive, limiting the number of patients who can receive timely radiotherapy. Additionally, CT scans offer lower soft-tissue contrast compared to MRI. Despite MRI providing superior soft-tissue visualization, its time-consuming nature makes it infeasible for real-time treatment planning. To address these challenges, we propose a method called \textbf{SegReg}, which utilizes Elastic Symmetric Normalization for registering MRI to perform OAR segmentation. SegReg outperforms the CT-only baseline by \textbf{16.78\%} in mDSC and \textbf{18.77\%} in mIoU, showing that it effectively combines the geometric accuracy of CT with the superior soft-tissue contrast of MRI, making accurate automated OAR segmentation for clinical practice become possible. See project website \textbf{\textcolor{blue}{\url{https://steve-zeyu-zhang.github.io/SegReg}}}
\end{abstract}

\vspace{-0.5cm}
\extrafootertext{$^{*}$Work done while being a visiting student researcher at Australian Institute for Machine Learning, The University of Adelaide.}
\section{Introduction}
\label{sec:intro}

The global incidence of head and neck cancer is on the rise with a projected increase of 30 percent annually by 2030 \cite{gormley2022reviewing}. Treatment modalities for head and neck cancer have evolved over time with the introduction of intensity modulated radiotherapy (IMRT) in the 1990s \cite{gujral2018patterns,nutting2023dysphagia,caudell2017future}. It is a valuable modality as there are important radiosensitive organs within close proximity of the target tissue \cite{eisbruch2002clinical}. 
IMRT is able to deliver highly conformal and homogenous radiation doses \cite{zelefsky2001high} to target tumours, while reducing dose in normal anatomical structures, i.e. organs at risk (OARs) \cite{sapkaroski2015review}. During radiotherapy (RT), precise control of radiation dose to OARs is essential to minimize post-treatment complications \cite{harari2010emphasizing}. Meanwhile, the dose within the planning target volume (PTV) should be tailored to achieve an optimal dose distribution \cite{inagaki2022escalated}.  
In image-guided radiotherapy (IGRT), this requires accurate segmentation of OARs in the radiotherapy computed tomography (RTCT) \cite{srinivasan2014applications} or the cone-beam computed tomography (CBCT) \cite{reggiori2011cone}, during radiotherapy treatment planning. 
In clinical practice, OAR segmentations are predominantly conducted manually by radiation oncologists, a process which is not only time-consuming, taking over 2 hours to segment nine OARs \cite{guo2020organ}, but also exhibits significant variability between different practitioners \cite{harari2010emphasizing}. Additionally, the wide size variations of OARs can make the process even more time-consuming to annotate compared to smaller structures. Predictably, as more OARs need to be included, the time requirements increase substantially, which in turn limits the number of patients who can receive timely radiotherapy \cite{mikeljevic2004trends}. These challenges have prompted efforts to develop automatic OAR segmentation methods for RT treatment planning.

\begin{figure}[h]
    \centering
    \includegraphics[width=\linewidth]{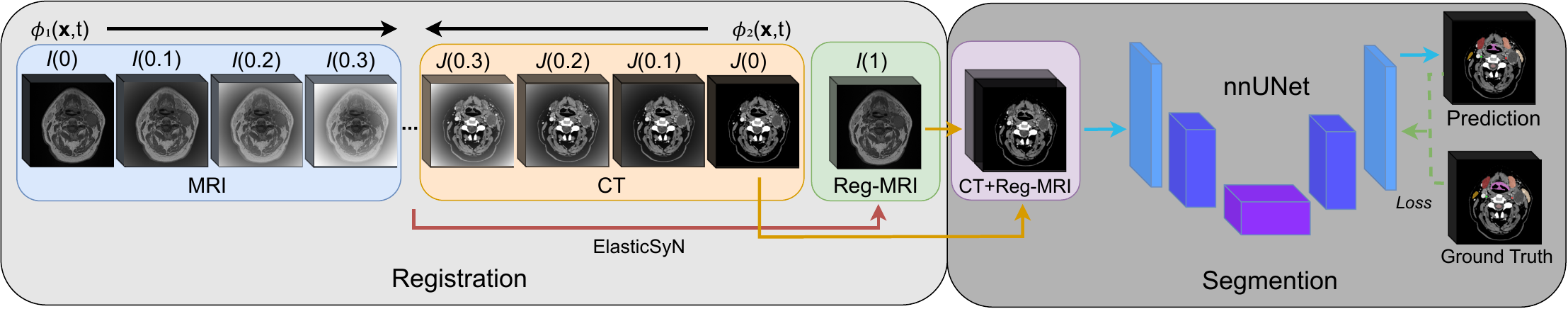} 
    \vspace{-0.8cm}
    \caption{The diagram illustrates the SegReg pipeline, which incorporates two main components: Elastic Symmetric Normalization (ElasticSyN) for aligning MRI to CT scans, and an nnU-Net model for segmenting Organs at Risk (OAR).}
    \label{fig:pipeline}
\end{figure}

While CT has traditionally served as the standard imaging modality for RT planning due to its geometric fidelity and the electron density (ED) information for dose calculations \cite{bahig2019clinical, johnstone2020guidance}, the inherently low image contrast of OARs in RTCT has been a limitation \cite{guo2020organ}. Over the past few decades, the integration of MRI into radiotherapy planning has become a standard practice in many clinical settings since it provides superior soft-tissue contrast compared to CT \cite{bahig2019clinical}. This adoption has facilitated more accurate OAR segmentation compared to CT \cite{savenije2020clinical}. Furthermore, it is now possible to plan treatments exclusively using MRI, without the need for RTCT \cite{johnstone2020guidance}. Given that MRI typically offers lower geometrical precision than CT and lacks inherent electron density information \cite{johnstone2020guidance}, dose calculations in such cases are performed by bulk electron density assignment \cite{davis2017can, prior2017bulk} or voxel-based techniques such as the use of synthetic CT (synCT) generated from MRI data \cite{johnstone2020guidance, prior2017bulk}. Nevertheless, the emergence of real-time MRI-guided radiotherapy, which often requires a significantly longer scan time compared to CT, along with the time-consuming manual OAR annotation, extends the total time for RT treatment planning \cite{savenije2020clinical}. 
This becomes a critical bottleneck in implementing MRI-only treatment planning in real-time adaptive radiotherapy \cite{savenije2020clinical, keall2019see}. 

In this paper, we present a simple yet effective pipeline known as \textbf{SegReg}, which harnesses co-registered MRI in conjunction with planning CT to perform multimodal OAR segmentation. This approach combines the superior soft-tissue contrast of MRI to enhance semantic knowledge and the high geometrical accuracy of CT to improve the shape of masks for OAR segmentation. This advancement pushes the boundaries of knowledge in OAR segmentation for IGRT in an automated fashion. It tackles the issue of low image contrast in OAR during CT-guided treatment planning and addresses the slowness associated with MR-guided planning, eliminating the need for patients to undergo time-consuming MRI scans in real-time during treatment planning. With its remarkable performance, this innovation holds the promise of widespread adoption in clinical practice.

\begin{figure}[H]
    \centering
    \includegraphics[width=\linewidth]{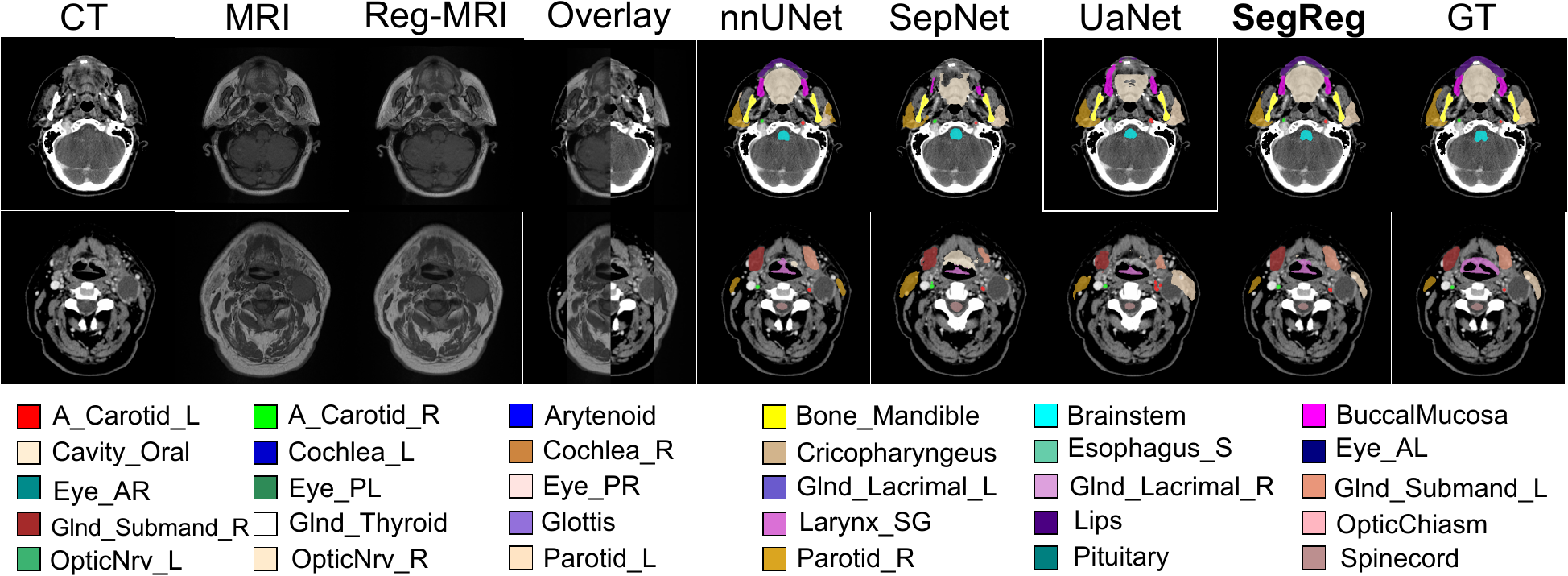} 
    \vspace{-0.8cm}
    \caption{The figure demonstrates visualizations of CT, MRI, and registration results, along with the comparison of OAR segmentation from both the proposed SegReg and other established methods. It demonstrates that SegReg outperforms the others in terms of both semantic accuracy and geometric fidelity.}
    \label{fig:demo}
    \vspace{-0.5cm}
\end{figure}

\section{Related Works}
\subsection{OARs Segmentation}

OAR segmentation has stood as a central research focus within the realm of RT treatment planning. Over time, several notable attempts have been made in this field. For instance, SOARS \cite{guo2020organ} introduced a technique that categorizes OARs into anchor, mid-level, and small \& hard groups, employing differentiable neural architecture search atop a fully convolutional network. UaNet \cite{tang2019clinically}, on the other hand, put forward an attention-modulated U-Net,
adopting a two-stage approach for OAR segmentation. The first stage involves detection, followed by segmentation. In a similar vein, SepNet \cite{lei2021automatic} presented a novel strategy that uses hard voxel weighting, leveraging a hardness-weighted loss. This approach places heightened emphasis on small organs and challenging voxels in larger, less complex organs. 
Each of these methods represents significant strides in advancing the field of OAR segmentation for improved RT treatment planning.

\vspace{-0.2cm}
\subsection{Image Registration}
\vspace{-0.1cm}

Image registration is a critical technique in medical imaging analysis, extensively employed in pathology, microscopy, surgical planning, and various other applications \cite{avants2014insight}. Numerous transformation algorithms have been utilized in clinical contexts, including B-Spline registration in Elastix \cite{klein2009elastix}, elastic-type models like HAMMER \cite{shen2002hammer}, and diffeomorphic algorithms such as DARTEL \cite{ashburner2007fast}. Each registration method possesses its own strengths and weaknesses \cite{avants2014insight}. For example, linear transformations like rigid transformation are often constrained by distortions in the images. In the context of multi-modalities, registration becomes more challenging as it involves aligning images from different acquisition techniques, such as CT and MRI. Challenges often arise when dealing with samples lying outside the region of interest in the moving image due to inadequate overlap between the input images.

\subsection{Registration Segmentation}

Multiple prior efforts have combined registered medical images with semantic segmentation techniques. For instance, ProRSeg \cite{jiang2023progressively} introduced a 3D convolutional recurrent registration approach to align MRI with cone-beam CT and then applied a recurrent segmentation network to segment OARs. Another notable work is HaN-Seg \cite{podobnik2023han}, which introduced a dataset featuring 30 OAR semantics and 42 pairs of CT and T1-weighted MRI training data. They proposed a baseline approach that utilized B-spline transformations with Elastix \cite{klein2009elastix} for MRI to CT registration and nnU-Net \cite{isensee2021nnu} backbone for segmentation. Additionally, the Modality Fusion Module (MFM) \cite{podobnik2023multimodal} emerged as an extension of the HaN-Seg baseline. MFM employed a double encoder architecture to separately encode CT and MRI information and reached a mDSC of 76.70\% in the HaN-Seg dataset using a 4-fold cross-validation without a hold-out test set. Similar to MFM, Modality-aware Mutual Learning (MAML) \cite{zhang2021modality} is also a two-stream early fusion network but using a mutual learning strategy composed of inter-intra joint loss. These prior attempts have significantly contributed to the field of registration segmentation, yet they have primarily treated registration as a technique without fully exploring its potential and conducting thorough ablation studies to reveal the intricacies of registration implementation.

\begin{table*}
\centering
\caption{
The table compares SegReg with the nnU-Net baseline for each semantic. It demonstrates that SegReg significantly outperforms nnU-Net, particularly for small tissues.} 
\label{table:3} 
\resizebox{1\textwidth}{!}{%
\begin{tabular}{c|c c c|c c c|c c c|c c c|c c c|c c c}
\toprule
\multicolumn{1}{c}{} & \multicolumn{3}{|c}{A\_Carotid\_L} & \multicolumn{3}{|c}{A\_Carotid\_R} & \multicolumn{3}{|c}{Arytenoid} & \multicolumn{3}{|c}{Bone\_Mandible} & \multicolumn{3}{|c}{Brainstem} & \multicolumn{3}{|c}{BuccalMucosa}\\ 
 & DSC $\uparrow$ & IoU $\uparrow$ & HD\textsubscript{95} $\downarrow$ & DSC $\uparrow$ & IoU $\uparrow$ & HD\textsubscript{95} $\downarrow$ & DSC $\uparrow$ & IoU $\uparrow$ & HD\textsubscript{95} $\downarrow$ & DSC $\uparrow$ & IoU $\uparrow$ & HD\textsubscript{95} $\downarrow$ & DSC $\uparrow$ & IoU $\uparrow$ & HD\textsubscript{95} $\downarrow$ & DSC $\uparrow$ & IoU $\uparrow$ & HD\textsubscript{95} $\downarrow$ \\ \midrule 
nnU-Net & 80.38 & 67.19 & 19.217 & 81.64 & 68.98 & 20.903 & \textbf{68.99} & \textbf{52.66} & \textbf{1.951} & \textbf{96.21} & \textbf{92.69} & \textbf{1.220} & 86.36 & 75.99 & 3.965 & \textbf{73.61} & \textbf{58.24} & \textbf{5.472} \\
\textbf{SegReg} & \textbf{86.77\textcolor{mygreen}{\raisebox{-0.8ex}{\textsuperscript{ +6.39}}}} & \textbf{76.63\textcolor{mygreen}{\raisebox{-0.8ex}{\textsuperscript{ +9.44}}}} & \textbf{5.054\textcolor{mygreen}{\raisebox{-0.8ex}{\textsuperscript{ -14.163}}}} & \textbf{87.46\textcolor{mygreen}{\raisebox{-0.8ex}{\textsuperscript{ +5.82}}}} & \textbf{77.71\textcolor{mygreen}{\raisebox{-0.8ex}{\textsuperscript{ +8.73}}}} & \textbf{1.120\textcolor{mygreen}{\raisebox{-0.8ex}{\textsuperscript{ -19.783}}}} & 68.58\textcolor{red}{\raisebox{-0.8ex}{\textsuperscript{ -0.41}}} & 52.19\textcolor{red}{\raisebox{-0.8ex}{\textsuperscript{ -0.47}}} & 2.052\textcolor{red}{\raisebox{-0.8ex}{\textsuperscript{ +0.101}}} & 96.01\textcolor{red}{\raisebox{-0.8ex}{\textsuperscript{ -0.20}}} & 92.33\textcolor{red}{\raisebox{-0.8ex}{\textsuperscript{ -0.36}}} & 1.580\textcolor{red}{\raisebox{-0.8ex}{\textsuperscript{ +0.36}}} & \textbf{89.58\textcolor{mygreen}{\raisebox{-0.8ex}{\textsuperscript{ +3.22}}}} & \textbf{81.13\textcolor{mygreen}{\raisebox{-0.8ex}{\textsuperscript{ +5.14}}}} & \textbf{3.718\textcolor{mygreen}{\raisebox{-0.8ex}{\textsuperscript{ -0.247}}}} & 73.58\textcolor{red}{\raisebox{-0.8ex}{\textsuperscript{ -0.03}}} & 58.20\textcolor{red}{\raisebox{-0.8ex}{\textsuperscript{ -0.04}}} & 5.746\textcolor{red}{\raisebox{-0.8ex}{\textsuperscript{ +0.274}}} \\
\midrule
\multicolumn{1}{c}{} & \multicolumn{3}{|c}{Cavity\_Oral} & \multicolumn{3}{|c}{Cochlea\_L} & \multicolumn{3}{|c}{Cochlea\_R} & \multicolumn{3}{|c}{Cricopharyngeus} & \multicolumn{3}{|c}{Esophagus\_S} & \multicolumn{3}{|c}{Eye\_AL}\\ 
 & DSC $\uparrow$ & IoU $\uparrow$ & HD\textsubscript{95} $\downarrow$ & DSC $\uparrow$ & IoU $\uparrow$ & HD\textsubscript{95} $\downarrow$ & DSC $\uparrow$ & IoU $\uparrow$ & HD\textsubscript{95} $\downarrow$ & DSC $\uparrow$ & IoU $\uparrow$ & HD\textsubscript{95} $\downarrow$ & DSC $\uparrow$ & IoU $\uparrow$ & HD\textsubscript{95} $\downarrow$ & DSC $\uparrow$ & IoU $\uparrow$ & HD\textsubscript{95} $\downarrow$ \\ \midrule 
nnU-Net & \textbf{91.54} & \textbf{84.39} & \textbf{4.507} & 55.11 & 38.04 & 51.452 & 54.66 & 37.61 & 18.345 & \textbf{71.14} & \textbf{55.21} & 4.317 & 53.42 & 36.44 & 8.010 & 48.67 & 32.16 & 41.052 \\
\textbf{SegReg} & 91.44\textcolor{red}{\raisebox{-0.8ex}{\textsuperscript{ -0.10}}} & 84.23\textcolor{red}{\raisebox{-0.8ex}{\textsuperscript{ -0.16}}} & \textbf{4.507} & \textbf{86.67\textcolor{mygreen}{\raisebox{-0.8ex}{\textsuperscript{ +31.56}}}} & \textbf{76.48\textcolor{mygreen}{\raisebox{-0.8ex}{\textsuperscript{ +38.44}}}} & \textbf{1.097\textcolor{mygreen}{\raisebox{-0.8ex}{\textsuperscript{ -50.355}}}} & \textbf{80.92\textcolor{mygreen}{\raisebox{-0.8ex}{\textsuperscript{ +26.26}}}} & \textbf{67.96\textcolor{mygreen}{\raisebox{-0.8ex}{\textsuperscript{ +30.35}}}} & \textbf{1.226\textcolor{mygreen}{\raisebox{-0.8ex}{\textsuperscript{ -17.119}}}} & 69.45\textcolor{red}{\raisebox{-0.8ex}{\textsuperscript{ -1.69}}} & 53.19\textcolor{red}{\raisebox{-0.8ex}{\textsuperscript{ -2.02}}} & \textbf{4.156\textcolor{mygreen}{\raisebox{-0.8ex}{\textsuperscript{ -0.161}}}} & \textbf{60.01\textcolor{mygreen}{\raisebox{-0.8ex}{\textsuperscript{ +6.59}}}} & \textbf{42.87\textcolor{mygreen}{\raisebox{-0.8ex}{\textsuperscript{ +6.43}}}} & \textbf{7.355\textcolor{mygreen}{\raisebox{-0.8ex}{\textsuperscript{ -0.655}}}} & \textbf{80.90\textcolor{mygreen}{\raisebox{-0.8ex}{\textsuperscript{ +32.23}}}} & \textbf{67.92\textcolor{mygreen}{\raisebox{-0.8ex}{\textsuperscript{ +35.76}}}} & \textbf{1.768\textcolor{mygreen}{\raisebox{-0.8ex}{\textsuperscript{ -39.284}}}} \\
\midrule
\multicolumn{1}{c}{} & \multicolumn{3}{|c}{Eye\_AR} & \multicolumn{3}{|c}{Eye\_PL} & \multicolumn{3}{|c}{Eye\_PR} & \multicolumn{3}{|c}{Glnd\_Lacrimal\_L} & \multicolumn{3}{|c}{Glnd\_Lacrimal\_R} & \multicolumn{3}{|c}{Glnd\_Submand\_L}\\ 
 & DSC $\uparrow$ & IoU $\uparrow$ & HD\textsubscript{95} $\downarrow$ & DSC $\uparrow$ & IoU $\uparrow$ & HD\textsubscript{95} $\downarrow$ & DSC $\uparrow$ & IoU $\uparrow$ & HD\textsubscript{95} $\downarrow$ & DSC $\uparrow$ & IoU $\uparrow$ & HD\textsubscript{95} $\downarrow$ & DSC $\uparrow$ & IoU $\uparrow$ & HD\textsubscript{95} $\downarrow$ & DSC $\uparrow$ & IoU $\uparrow$ & HD\textsubscript{95} $\downarrow$ \\ \midrule 
nnU-Net & 36.76 & 22.52 & 33.925 & 45.51 & 29.46 & 35.241 & 35.74 & 21.76 & 29.087 & 32.23 & 19.21 & 58.022 & 23.66 & 13.42 & 37.181 & 81.36 & 68.58 & 3.991 \\
\textbf{SegReg} & \textbf{81.13\textcolor{mygreen}{\raisebox{-0.8ex}{\textsuperscript{ +44.37}}}} & \textbf{68.26\textcolor{mygreen}{\raisebox{-0.8ex}{\textsuperscript{ +45.74}}}} & \textbf{1.797\textcolor{mygreen}{\raisebox{-0.8ex}{\textsuperscript{ -32.128}}}} & \textbf{94.49\textcolor{mygreen}{\raisebox{-0.8ex}{\textsuperscript{ +48.98}}}} & \textbf{89.56\textcolor{mygreen}{\raisebox{-0.8ex}{\textsuperscript{ +60.10}}}} & \textbf{1.256\textcolor{mygreen}{\raisebox{-0.8ex}{\textsuperscript{ -33.985}}}} & \textbf{93.82\textcolor{mygreen}{\raisebox{-0.8ex}{\textsuperscript{ +58.08}}}} & \textbf{88.36\textcolor{mygreen}{\raisebox{-0.8ex}{\textsuperscript{ +66.6}}}} & \textbf{1.515\textcolor{mygreen}{\raisebox{-0.8ex}{\textsuperscript{ -27.572}}}} & \textbf{67.07\textcolor{mygreen}{\raisebox{-0.8ex}{\textsuperscript{ +34.84}}}} & \textbf{50.45\textcolor{mygreen}{\raisebox{-0.8ex}{\textsuperscript{ +31.24}}}} & \textbf{12.642\textcolor{mygreen}{\raisebox{-0.8ex}{\textsuperscript{ -45.380}}}} & \textbf{70.00\textcolor{mygreen}{\raisebox{-0.8ex}{\textsuperscript{ +46.34}}}} & \textbf{53.84\textcolor{mygreen}{\raisebox{-0.8ex}{\textsuperscript{ +40.42}}}} & \textbf{2.185\textcolor{mygreen}{\raisebox{-0.8ex}{\textsuperscript{ -34.996}}}} & \textbf{87.08\textcolor{mygreen}{\raisebox{-0.8ex}{\textsuperscript{ +5.72}}}} & \textbf{77.12\textcolor{mygreen}{\raisebox{-0.8ex}{\textsuperscript{ +8.54}}}} & \textbf{3.076\textcolor{mygreen}{\raisebox{-0.8ex}{\textsuperscript{ -0.915}}}} \\
\midrule
\multicolumn{1}{c}{} & \multicolumn{3}{|c}{Glnd\_Submand\_R} & \multicolumn{3}{|c}{Glnd\_Thyroid} & \multicolumn{3}{|c}{Glottis} & \multicolumn{3}{|c}{Larynx\_SG} & \multicolumn{3}{|c}{Lips} & \multicolumn{3}{|c}{OpticChiasm}\\ 
 & DSC $\uparrow$ & IoU $\uparrow$ & HD\textsubscript{95} $\downarrow$ & DSC $\uparrow$ & IoU $\uparrow$ & HD\textsubscript{95} $\downarrow$ & DSC $\uparrow$ & IoU $\uparrow$ & HD\textsubscript{95} $\downarrow$ & DSC $\uparrow$ & IoU $\uparrow$ & HD\textsubscript{95} $\downarrow$ & DSC $\uparrow$ & IoU $\uparrow$ & HD\textsubscript{95} $\downarrow$ & DSC $\uparrow$ & IoU $\uparrow$ & HD\textsubscript{95} $\downarrow$ \\ \midrule 
nnU-Net & 85.49 & 74.65 & 24.122 & 90.04 & 81.88 & 2.192 & 77.33 & 63.03 & 2.503 & 83.05 & 71.01 & 3.505 & 78.77 & 64.97 & 5.248 & 42.94 & 27.34 & 2.937 \\
\textbf{SegReg} & \textbf{89.77\textcolor{mygreen}{\raisebox{-0.8ex}{\textsuperscript{ +4.28}}}} & \textbf{81.44\textcolor{mygreen}{\raisebox{-0.8ex}{\textsuperscript{ +6.79}}}} & \textbf{2.020\textcolor{mygreen}{\raisebox{-0.8ex}{\textsuperscript{ -22.102}}}} & \textbf{91.02\textcolor{mygreen}{\raisebox{-0.8ex}{\textsuperscript{ +0.98}}}} & \textbf{83.52\textcolor{mygreen}{\raisebox{-0.8ex}{\textsuperscript{ +1.64}}}} & \textbf{1.876\textcolor{mygreen}{\raisebox{-0.8ex}{\textsuperscript{ -0.316}}}} & \textbf{79.28\textcolor{mygreen}{\raisebox{-0.8ex}{\textsuperscript{ +1.95}}}} & \textbf{65.67\textcolor{mygreen}{\raisebox{-0.8ex}{\textsuperscript{ +2.64}}}} & \textbf{2.261\textcolor{mygreen}{\raisebox{-0.8ex}{\textsuperscript{ -0.242}}}} & \textbf{84.27\textcolor{mygreen}{\raisebox{-0.8ex}{\textsuperscript{ +1.22}}}} & \textbf{72.81\textcolor{mygreen}{\raisebox{-0.8ex}{\textsuperscript{ +1.8}}}} & \textbf{3.435\textcolor{mygreen}{\raisebox{-0.8ex}{\textsuperscript{ -0.07}}}} & \textbf{81.04\textcolor{mygreen}{\raisebox{-0.8ex}{\textsuperscript{ +2.27}}}} & \textbf{68.12\textcolor{mygreen}{\raisebox{-0.8ex}{\textsuperscript{ +3.15}}}} & \textbf{4.686\textcolor{mygreen}{\raisebox{-0.8ex}{\textsuperscript{ -0.562}}}} & \textbf{50.56\textcolor{mygreen}{\raisebox{-0.8ex}{\textsuperscript{ +7.62}}}} & \textbf{33.83\textcolor{mygreen}{\raisebox{-0.8ex}{\textsuperscript{ +6.49}}}} & \textbf{2.242\textcolor{mygreen}{\raisebox{-0.8ex}{\textsuperscript{ -0.695}}}} \\
\midrule
\multicolumn{1}{c}{} & \multicolumn{3}{|c}{OpticNrv\_L} & \multicolumn{3}{|c}{OpticNrv\_R} & \multicolumn{3}{|c}{Parotid\_L} & \multicolumn{3}{|c}{Parotid\_R} & \multicolumn{3}{|c}{Pituitary} & \multicolumn{3}{|c}{SpinalCord}\\ 
 & DSC $\uparrow$ & IoU $\uparrow$ & HD\textsubscript{95} $\downarrow$ & DSC $\uparrow$ & IoU $\uparrow$ & HD\textsubscript{95} $\downarrow$ & DSC $\uparrow$ & IoU $\uparrow$ & HD\textsubscript{95} $\downarrow$ & DSC $\uparrow$ & IoU $\uparrow$ & HD\textsubscript{95} $\downarrow$ & DSC $\uparrow$ & IoU $\uparrow$ & HD\textsubscript{95} $\downarrow$ & DSC $\uparrow$ & IoU $\uparrow$ & HD\textsubscript{95} $\downarrow$ \\ \midrule 
nnU-Net & 40.82 & 25.64 & 21.363 & 47.23 & 30.92 & 20.852 & 47.76 & 31.37 & 61.995 & 63.18 & 46.18 & 58.114 & 75.83 & 61.07 & 1.816 & 84.91 & 73.78 & 2.335\\
\textbf{SegReg} & \textbf{79.76\textcolor{mygreen}{\raisebox{-0.8ex}{\textsuperscript{ +38.94}}}} & \textbf{66.33\textcolor{mygreen}{\raisebox{-0.8ex}{\textsuperscript{ +40.69}}}} & \textbf{1.767\textcolor{mygreen}{\raisebox{-0.8ex}{\textsuperscript{ -19.596}}}} & \textbf{75.59\textcolor{mygreen}{\raisebox{-0.8ex}{\textsuperscript{ +28.36}}}} & \textbf{60.75\textcolor{mygreen}{\raisebox{-0.8ex}{\textsuperscript{ +29.83}}}} & \textbf{1.815\textcolor{mygreen}{\raisebox{-0.8ex}{\textsuperscript{ -19.037}}}} & \textbf{88.37\textcolor{mygreen}{\raisebox{-0.8ex}{\textsuperscript{ +40.61}}}} & \textbf{79.16\textcolor{mygreen}{\raisebox{-0.8ex}{\textsuperscript{ +47.79}}}} & \textbf{15.805\textcolor{mygreen}{\raisebox{-0.8ex}{\textsuperscript{ -46.19}}}} & \textbf{88.64\textcolor{mygreen}{\raisebox{-0.8ex}{\textsuperscript{ +25.46}}}} & \textbf{79.59\textcolor{mygreen}{\raisebox{-0.8ex}{\textsuperscript{ +33.41}}}} & \textbf{5.100\textcolor{mygreen}{\raisebox{-0.8ex}{\textsuperscript{ -53.014}}}} & \textbf{78.95\textcolor{mygreen}{\raisebox{-0.8ex}{\textsuperscript{ +3.12}}}} & \textbf{65.23\textcolor{mygreen}{\raisebox{-0.8ex}{\textsuperscript{ +4.16}}}} & \textbf{1.535\textcolor{mygreen}{\raisebox{-0.8ex}{\textsuperscript{ -0.281}}}} & \textbf{85.55\textcolor{mygreen}{\raisebox{-0.8ex}{\textsuperscript{ +0.64}}}} & \textbf{74.75\textcolor{mygreen}{\raisebox{-0.8ex}{\textsuperscript{ +0.97}}}} & \textbf{2.198\textcolor{mygreen}{\raisebox{-0.8ex}{\textsuperscript{ -0.137}}}} \\
\midrule
\end{tabular}%
}
\end{table*}

\section{Methodology}
\vspace{-0.2cm}

The SegReg involves two stages: an Elastic Symmetric Normalization (ElasticSyN) transformation \cite{avants2008symmetric} for registering MRI to CT and an nnU-Net model for OAR segmentation \cite{isensee2021nnu}, which shown in figure \ref{fig:pipeline}.

During the registration process, within every pair of computed tomography (CT) and magnetic resonance imaging (MRI), the moving image (MRI) is aligned with the fixed image (CT) utilizing an Elastic Symmetric Normalization (ElasticSyN) transformation, yielding a registered MRI (Reg-MRI).

\vspace{-0.4cm}
\begin{equation}
    \text{Reg-MRI} = ElasticSyN(\text{MRI}, \text{CT})
\end{equation}

The registered MRI will be combined with CT into a two-channel volume as input $X$. The nnU-Net utilizes the ground truth $Y$ for supervision to train network $T$ using a combined loss function of weighted cross-entropy loss and weighted Dice loss.

\vspace{-0.6cm}
\begin{equation}
    L = W_{\text{CE}} l_{\text{CE}}(T(X),Y) + W_{\text{Dice}} l_{\text{Dice}}(T(X),Y) 
\end{equation}

\section{Experiments}

\begin{table}[H]
\centering
\resizebox{0.4\textwidth}{!}{%
\begin{tabular}{c|cccc}
\toprule
Models & mDSC & aDSC & mIoU & aIoU  \\ \midrule 
nnU-Net & 64.48 & 89.07 & 50.88 & 80.29 \\
SepNet & 65.46 & 76.50 & 52.39 & 61.94 \\
UaNet & 67.57 & 75.45 & 53.06 & 60.57 \\
\textbf{SegReg (Ours)} & \textbf{81.26} & \textbf{89.75} & \textbf{69.65} & \textbf{81.40} \\
\midrule
\end{tabular}%
}
\caption{The table compares the proposed SegReg model with other established OAR segmentation models. It demonstrates that SegReg achieves state-of-the-art performance.} 
\label{table:1} 
\end{table}

\subsection{Experiment Setup}

We performed our experiments using the HaN-Seg \cite{podobnik2023han} dataset, comprising 42 pairs of CT and T1-weighted public scans with pixel-level annotations across 30 distinct OARs. We randomly split the dataset into a training set with 38 instances and an evaluation set with 4 instances, and the training set is trained in 5-fold cross validation. For the baseline, we trained a vanilla nnU-Net \cite{isensee2021nnu} on 38 CT scans. Additionally, for comparative purposes, we applied several OAR segmentation models, including SepNet \cite{lei2021automatic}, and UaNet \cite{tang2019clinically}
to the same set of CT scans. Next, we trained the proposed SegReg using the paired CT and T1-weighted MRI from the training set. Subsequently, we evaluated the performance of these models on the 4 test instances.

We assessed the model's performance using several evaluation metrics: mean Dice Similarity Coefficient (mDSC) and mean Intersection over Union (mIoU) to gauge overall performance across semantic categories, and class-agnostic Dice Similarity Coefficient (aDSC) and class-agnostic Intersection over Union (aIoU) to evaluate segmentation shape by treating all semantics as a single foreground semantic. Additionally, we employed the $95^{th}$-percentile Hausdorff distance (HD\textsubscript{95}) to account for outliers, as recommended in prior studies \cite{vrtovec2020auto, maier2020bias, nikolov2021clinically}, making it particularly suitable for assessing small volumetric structures and results aligned with interrater variability.

\begin{table}[H]
\centering
\resizebox{0.15\textwidth}{!}{%
\begin{tabular}{c|c}
\toprule
Models & mDSC   \\ \midrule 
MAML & 76.30 \\
MFM & 76.70 \\
\textbf{SegReg} & \textbf{80.29} \\
\midrule
\end{tabular}%
}
\caption{The table shows the comparsion of proposed SegReg with latest modality fusion models, demonstrating that SegReg outperforms two-stream networks.} 
\label{table:2} 
\end{table}

\subsection{Results}

The results compared with nnU-Net baseline, including a detailed breakdown for each semantic, are presented in Table \ref{table:3}. Our model demonstrates a notable performance improvement, especially in small and tiny organs, including the cochlea, anterior/posterior eyeball, lacrimal gland, optic nerves, and parotid gland. Furthermore, the comparative results presented in Table \ref{table:1} and Figure \ref{fig:demo} highlight a notable improvement in our model's performance when compared to the CT-only baseline and other established OAR segmentation models. Specifically, in terms of semantic classification ability, our model has achieved a 16.78\% improvement in mDSC compared to the nnU-Net baseline. In addition, when considering models that also incorporate MRI data, it's worth noting that both MAML \cite{zhang2021modality} and MFM \cite{podobnik2023multimodal} employ a four-fold cross-validation approach on the entire dataset without a separate hold-out test set, we have also conducted experiments with our SegReg model under the same setting. The results presented in Table \ref{table:2} demonstrate that our SegReg model continues to outperform two-stream networks, irrespective of the modality fusion architectures used in multi-modal OAR segmentation.

\begin{table}[H]
\centering
\resizebox{0.5\textwidth}{!}{%
\begin{tabular}{c|cccc}
\toprule
Models & mDSC & aDSC & mIoU & aIoU   \\ \midrule 
CT & 64.48 & 89.07 & 50.88 & 80.29 \\
MRI\textsubscript{Reg} & 68.03\textcolor{mygreen}{\raisebox{-0.8ex}{\textsuperscript{ +3.55}}} & 83.95\textcolor{red}{\raisebox{-0.8ex}{\textsuperscript{ -5.12}}} & 53.34\textcolor{mygreen}{\raisebox{-0.8ex}{\textsuperscript{ +2.46}}} & 72.34\textcolor{red}{\raisebox{-0.8ex}{\textsuperscript{ -7.95}}} \\
\textbf{CT+MRI\textsubscript{Reg}} & \textbf{81.26\textcolor{mygreen}{\raisebox{-0.8ex}{\textsuperscript{ +16.78}}}} & \textbf{89.75\textcolor{mygreen}{\raisebox{-0.8ex}{\textsuperscript{ +0.68}}}} & \textbf{69.65\textcolor{mygreen}{\raisebox{-0.8ex}{\textsuperscript{ +18.77}}}} & \textbf{81.40\textcolor{mygreen}{\raisebox{-0.8ex}{\textsuperscript{ +1.11}}}}\\
\midrule
\end{tabular}%
}
\caption{The table demonstrates that utilizing registered MRI leads to superior performance in semantic recognition than CT-only baseline, attributable to the heightened soft-tissue contrast inherent in MRI imaging.} 
\label{table:4} 
\end{table}

\vspace{-0.6cm}

\begin{table}[H]
\centering
\resizebox{0.4\textwidth}{!}{%
\begin{tabular}{c|cccc}
\toprule
Models & mDSC & aDSC & mIoU & aIoU  \\ \midrule 
CT & 64.48 & 89.07 & 50.88 & 80.29 \\
+ Translation & 78.38 & 89.45 & 65.76 & 80.92 \\
+ Rigid & 80.29 & 89.52 & 68.19 & 81.03 \\
+ Affine & 79.56 & 89.51 & 67.51 & 81.01 \\
+ Elastic & 78.15 & 89.35 & 65.45 & 80.76 \\
\textbf{+ ElasticSyN} & \textbf{81.26} & \textbf{89.75} & \textbf{69.65} & \textbf{81.40} \\
\midrule
\end{tabular}%
}
\caption{The table illustrates the ablations of various transformations, indicating that ElasticSyN used in SegReg consistently outperforms other methods.} 
\label{table:5} 
\end{table}

\subsection{Ablation Studies}

To assess the extent of improvement achieved by the proposed registered MRI in OAR segmentation, we carried out an experiment focusing solely on the registered MRI data. The results in Table \ref{table:4} indicate that even when only using registered MRI, we achieve better performance in semantic classification, with an improvement of 3.55\% in mDCS. Despite the geometric fidelity of registered MRI not matching that of the originally annotated CT scans, which has shown in agnostic metrics, the improvement in semantics still demonstrates MRI offers superior localization and contrast in soft tissue for segmentation models compared with CT scans, making it more precisely to distinguish and delineate different OARs. Furthermore, combining the original CT scans with registered MRI leverages the superior soft-tissue contrast of MRI to enhance semantic knowledge, while benefiting from the high geometrical accuracy of CT to improve mask shapes for OAR segmentation. This results in an improvement of 16.78\% in mDSC and 18.77\% in mIoU.

We also explore various transformation components of the MRI registration in SegReg, including Translation, Rigid transformation (translation and rotation), Affine transformation (translation, rotation, and scaling), and Elastic transformation (affine and deformable transformation) \cite{avants2011reproducible}, in comparison to the Elastic Symmetric Normalization \cite{avants2008symmetric}. The results in Table \ref{table:5} indicate that Elastic Symmetric Normalization, as employed in SegReg, outperforms any other registration method in OAR segmentation.

Furthermore, we investigated the impact of a two-stream backbone on multi-modal OAR segmentation, comparing it to the vanilla single-stream network. We replaced the nnU-Net \cite{isensee2021nnu} backbone with the MAML \cite{zhang2021modality} backbone in SegReg, and the performance is presented in Table \ref{table:6}. The results indicate that the two-stream architecture has minimal impact compared to the significant contribution of registration transformation to overall OAR segmentation performance. Using a single-stream backbone remains a simple yet effective approach for registration segmentation.

\begin{table}[H]
\centering
\resizebox{0.45\textwidth}{!}{%
\begin{tabular}{c|cccc}
\toprule
Models & mDSC & aDSC & mIoU & aIoU  \\ \midrule 
nnU-Net & 64.48 & 89.07 & 50.88 & 80.29 \\
SegReg (MAML) & 78.94 & 89.11 & 66.61 & 80.37 \\
\textbf{SegReg (nnU-Net)} & \textbf{81.26} & \textbf{89.75} & \textbf{69.65} & \textbf{81.40} \\
\midrule
\end{tabular}%
}
\caption{The table provides a comparative analysis between the single stream backbone (nnU-Net) and the double stream backbone (MAML) concerning existing registration methodologies. The findings demonstrate superior performance of the single stream network.} 
\label{table:6} 
\end{table}

\section{Discussion and Conclusion}

In conclusion, SegReg significantly outperforms other renowned OAR segmentation models and effectively combines the geometric accuracy of CT with the superior soft-tissue contrast of MRI. Notably, SegReg excels in the segmentation of small organs, particularly eye-related tissues such as the anterior/posterior eyeballs, lacrimal glands, and optic nerves. This is crucial given that radiation-induced ocular complications are major side effects of radiation therapy, encompassing acute lesions in the eyelid, conjunctiva, and corneal epithelium, as well as delayed effects like cataracts, glaucoma, and retinopathy \cite{nuzzi2020ocular}. The notable improvement in the safety and practicality of automated OAR segmentation in clinical applications makes its use in clinical practice feasible.



\clearpage
{\fontsize{8.45}{8}\selectfont 
}

\end{document}